\documentclass{article}

\usepackage{PRIMEarxiv}

\usepackage[utf8]{inputenc} 
\usepackage[T1]{fontenc}    
\usepackage{hyperref}       
\usepackage{url}            
\usepackage{booktabs}       
\usepackage{amsfonts}       
\usepackage{nicefrac}       
\usepackage{microtype}      
\usepackage{lipsum}
\usepackage{fancyhdr}       
\usepackage{graphicx}       
\graphicspath{{media/}}     
\usepackage{amsmath}

\title{Anchoring Bias in Large Language Models: An Experimental Study
\thanks{This work has been submitted to the IEEE for possible publication. Copyright may be transferred without notice, after which this version may no longer be accessible.} 
}

\author{
  \uppercase{Jiaxu Lou ~~~~~~~~~~~~~ Yifan Sun} \\
  National Day School, Beijing, China \\
  \texttt{jiaxulou@hotmail.com; sunyifan@bnds.cn}
}

\begin{document}
\maketitle

\begin{abstract}
Large Language Models (LLMs) like GPT-4 and Gemini have significantly advanced artificial intelligence by enabling machines to generate and comprehend human-like text.
Despite their impressive capabilities, LLMs are not free of limitations. 
They have shown  various biases. 
While much research has explored demographic biases, the cognitive biases in LLMs have not been equally studied. 
This study delves into anchoring bias, a cognitive bias where initial information disproportionately influences judgment. 
Utilizing an experimental dataset, we examine how anchoring bias manifests in LLMs and verify the effectiveness of various mitigation strategies. 
Our findings highlight the sensitivity of LLM responses to biased hints.
At the same time, our experiments show that, to mitigate anchoring bias, one needs to collect hints from comprehensive angles to prevent the LLMs from being anchored to individual pieces of information, while simple algorithms such as Chain-of-Thought, Thoughts of Principles, Ignoring Anchor Hints, and Reflection are not  sufficient.
\end{abstract}

\section{Introduction}
\label{sec:introduction}
Nowadays, large language models (LLMs) such as GPT4 and Gemini have revolutionized the field of artificial intelligence by enabling machines to understand and generate human-like text. 
Furthermore, they also have shown remarkable performance in many commonsense reasoning tasks. 
Many researchers and industry practitioners have started applying LLMs into decision-making processes to support their day-to-day business. 
For example, besides being employed to generate articles, stories, and even code, LLMs are also used as chatbots and virtual assistants for instant 
support in customer service, to assist with medical research, diagnostics, and patient communication in healthcare \cite{Shojaei2024, Wei2024}, and many other decision-making scenarios.
Despite their impressive capabilities, LLMs are not free of limitations. In fact, a set of studies have shown that LLMs have social or stereo bias problems \cite{Yeh2023}, and these biases may bring negative impacts. 
A lot of attention has been paid to better understand and eliminate the 
negative impact of demographic bias in LLMs, by both the research community and industry.

Different from demographic biases of LLMs based on sensitive characteristics such as race and gender that have been widely studied by researchers \cite{Gallegos2024}, cognitive biases of LLMs have not attracted a lot attention yet. 
There are very few research papers focusing on cognitive biases in LLMs. 
As we know, cognitive biases in LLMs can affect the fairness and accuracy of their outputs \cite{Wang2024, Schmidgall2024}. 
It is especially important when we use LLMs in decision-making tasks, as it may lead to unreasonable recommendations.
As LLMs are more and more popular to be used in decision making scenarios, there is a high demand to have a comprehensive understanding of LLMs’ cognitive biases and let practitioners to be cautious about them to avoid irrational decision-making output.

This paper aims to focus on a specific type of cognitive biases – anchoring bias and study its manifest in LLMs and possible mitigation strategies.
Different from \cite{Jeremy2024} where only a few financial questions are examined, we conducted a more comprehensive and quantitative study on the anchoring bias of LLMs based on the experimental dataset designed by Taha Yasseri \cite{Yasseri2022}.
By changing the magnitude of bias hints, we found that the answers of LLMs are sensitive to the biased hints.
We further conducted a comparative study on several mitigation strategies. 
Our study shows that simple algorithms such as Chain-of-Thought, Thoughts of Principles, Ignoring Anchor Hints, and Reflection are not insufficient to mitigate anchoring bias.
One needs to collect hints from comprehensive angles to prevent the LLMs from being anchored to individual pieces of information.
\section{Related Work}
\label{sec:related_work}
\subsection{Cognitive Bias}
Cognitive bias refers to systematic mental patterns that influence our thinking and decision-making, leading us to process information in a selective and subjective manner, often resulting in inaccurate or irrational judgments \cite{Wilke2012}.
It is a systematic error in thinking, affecting how we process information, perceive the environment, and make decisions. 
It can lead to irrational thoughts or judgments and is often based on our perceptions, memories, or individual and societal beliefs.
Cognitive bias is often an unconscious and automatic process -- a result of  the brain’s attempt to simplify information processing that are designed to make decision-making quicker and more efficient.
Individuals are often unaware of  their attitudes and behaviors resulting from them \cite{Wilke2012}. 
Different from demographic bias where behavior occurs due to social and cultural influences, cognitive bias lies in the information processing mechanisms of human decision-making  process, often influenced by the setup of the task or emotional factors. 

There are a lot of cognitive bias patterns that have been studied in Psychology.
For example, confirmation bias, hindsight bias, mere exposure effect, self-serving bias, base rate fallacy, anchoring bias, availability bias, the framing effect, inattentional blindness, and the ecological fallacy are some of the most common examples of cognitive bias \cite{Wilke2012}. 

In this study, we mainly focus on anchoring bias \cite{Furnham2011}. It  means that our judgments are often significantly influenced by the first piece of information encountered.
For instance, once an anchor (i.e., the first piece of information encountered) is established from a previous message, people often insufficiently adjust away from it to arrive at their final answer, and so their final guess or decision is closer to the anchor than it otherwise would have been. 

\subsection{Cognitive Biases in LLMs}
LLMs are trained by a next-token prediction task based on the text data collected from the Internet. 
Therefore, LLMs are susceptible to various biases because the data they are trained on inherently includes biases. 
Additionally, the algorithms used to train these models can exacerbate or introduce new biases depending on their design and the assumptions they make during learning.
The interaction between users and the model can further entrench these biases, as frequent user inputs may reinforce certain patterns that the model learns to replicate.
These biases can potentially manifest in several ways at the LLM inference time, impacting the reliability and fairness of the model’s outputs. 

A few existing research studies have shown that LLMs can show cognitive biases in many different decision-making processes \cite{Chen2024, Echterhoff2024, Yeh2023}. 
For example, Chen et.al. \cite{Chen2024} pointed out that LLM’s judgments are influenced by threshold priming biases. 
In \cite{Echterhoff2024}, the authors detected several cognitive biases (confirmation bias, anchoring bias, status quo bias, framing bias, primacy bias, group attribution bias, etc. ) in LLMs by using a framework called BIASBUSTER. Samuel et.al.\cite{Schmidgall2024} assessed cognitive biases in LLMs used for medical tasks by using a benchmark called BiasMedQA, and showed that LLMs are vulnerable to cognitive biases,  while GPT-4 is demonstrating significant resistance to bias. 
However, the research of \cite{Olivia2024, Pantana2024} showed that although GPT models are frequently subject to cognitive biases, the way these biases are displayed does not reflect that shown by humans.
Ross et.al \cite{Ross2024} also evaluate the cognitive biases of LLMs quantitively using a utility theory in the financial domain and found that the financial behavior of current LLMs is neither entirely human-like nor entirely homo-economicus-like. 

In this paper, we focus on the anchoring bias of LLMs including its behavior and mitigation strategies. 
There are a few existing works focusing on anchoring bias of LLMs. 
Li et.al. \cite{Li2024} studied LLMs’ anchoring bias on answering multiple-choice questions (MCQs). 
To mitigate it, the authors identified the Multi-Layer Perceptron (MLP) and attention neural vectors that are responsible for this bias and updated these vectors to neutralize the bias to the anchored choice ‘A’. However, their approach can only be applied to limited scenarios  i.e., 
MCQs. 
The most related work to our research in this paper is from \cite{Jeremy2024} where the authors tested whether LLMs are subject to anchoring bias and studied two naïve mitigation prompting strategies.
They found that two naïve mitigation prompting strategies including Chain 
of Thought (CoT) and an “ignore previous” strategy (using a prompt: “Ignore our previous conversation but remember your role”) cannot consistently reduce biases for different LLMs. 
However, this study is only based on 5 questions in the financial domain. Different from these existing ones, our study includes 62 questions from different domains and gives a quantitative analysis about the effects of anchoring bias with different hints.
\section{Our Approach}
\label{sec:our_approach}
In this study, we conduct a systematic experiment to assess anchoring bias of LLMs through an in
depth quantitative analysis and explore potential mitigation strategies based on the dataset from 
\cite{Yasseri2022}. 

\begin{figure*}
    \centering
    \includegraphics[width=0.9\linewidth]{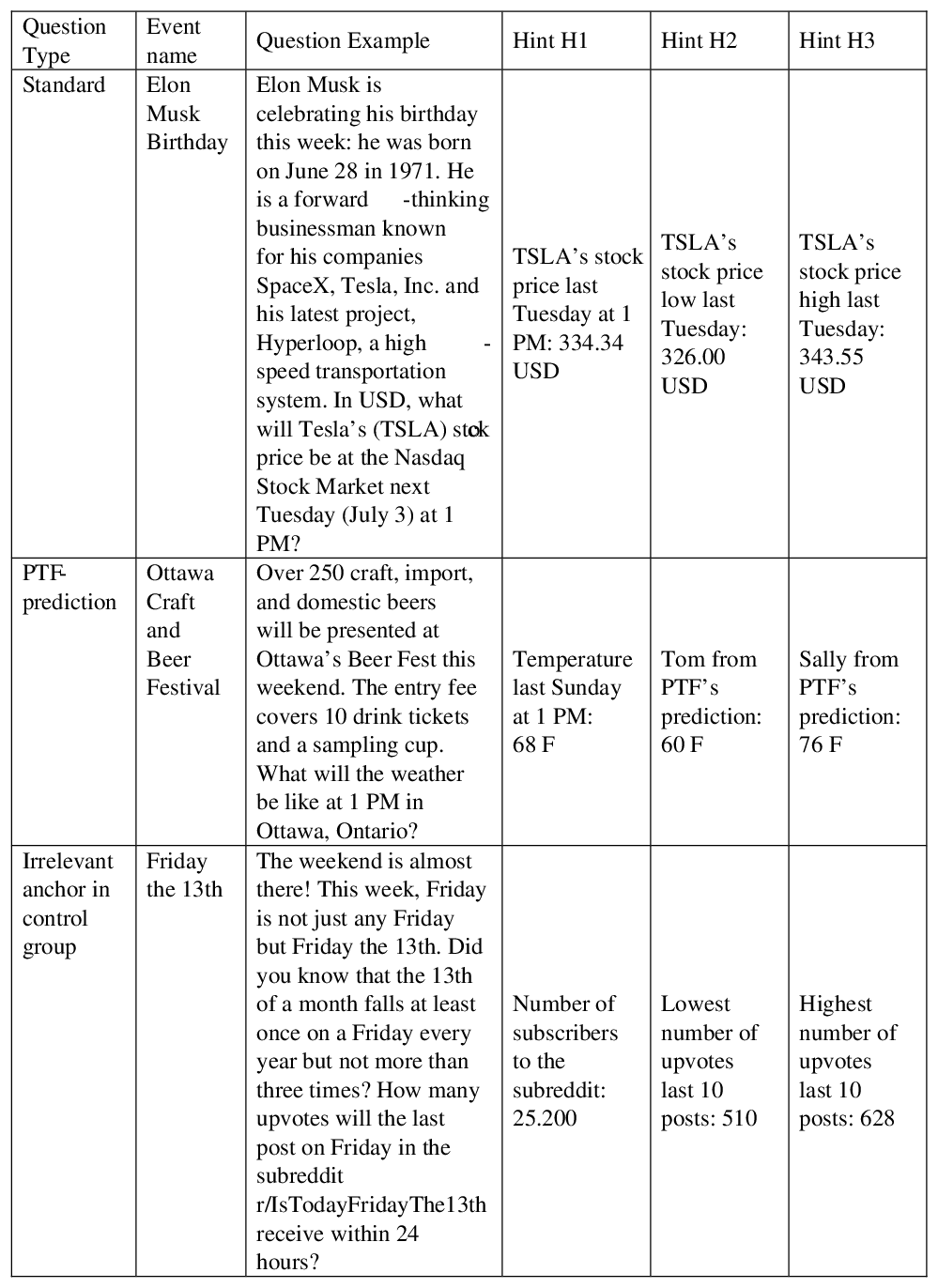}
    \caption{Question Examples}
    \label{fig:table1}
\end{figure*}

\subsection{Data Set}
In order to systematically evaluate the anchoring effects of LMMs, we use a dataset from \cite{Yasseri2022} which was originally collected for a quantitative analysis of human anchoring biases. 
To make the data representative and closer to everyday life, the data is collected from the community of the "Play the Future" (PTF) game, where users make digital predictions about the outcomes of various economic, social, sports, entertainment, and other events. 
Figure~\ref{fig:table1} gives three typical examples.
Each sample contains a question $Q$ requiring a numerical answer, and three hint messages $H_1$, $H_2$, and $H_3$. 
For example, “Over 250 craft, import, and domestic beers will be presented at Ottawa's Beer Fest this weekend. The entry fee covers 10 drink tickets and a sampling cup. What will the weather be like at 1 PM in Ottawa, Ontario, on Sunday?” is the question of the second sample in Figure~\ref{fig:table1}, which is asking for a temperature value. 
The first hint $H_1$ of this sample (“Temperature last Sunday at 1 PM: 68°F”) is a fact that usually  provides a useful reference information for answering the numerical question $Q$ of this sample.
$H_2$ and $H_3$ are two different anchoring hints that may potentially trigger the LLMs to give biased answers about the question topic.
In this case, $H_2$ ("Tom from PTF's prediction: 60 F") gives a low\-value anchor (i.e. 60 F) and $H_3$ gives a high\-value anchor (i.e. 76 F).
The whole dataset contains 62 samples (and their corresponding hints) covering multiple aspects of daily life, including some natural, economic, social, sports, and entertainment events, such as weather, stock prices, and flight times.

There are three different types of anchoring hints included in the dataset to assess different types of anchoring situations: fact anchoring,  “expert-opinion” anchoring, irrelevant anchoring.
Different types of anchoring questions may trigger different bias behaviors of LLMs. 
These different types of questions and hints allow us to evaluate how LLMs predictions are influenced by these different anchoring facts.
Figure~\ref{fig:table1} shows three typical question samples.
In the first row of Figure~\ref{fig:table1}, the hints of $H_2$ and $H_3$ demonstrate a fact anchoring situation, where the $H_2$ and $H_3$ anchoring hints are factual descriptions of the highest and lowest stock prices last Tuesday.
Although neither prompt can directly provide the answer to the question (next Tuesday's stock price at 1 PM), seeing different $H_2$ and $H_3$ prompts may lead to different LLM answers. 
In other words, these two hints may cause the models to show anchoring effects.
In the second row of Figure~\ref{fig:table1}, the $H_2$ and $H_3$ hints provide two different expert opinions, which we can use to evaluate the impact of expert opinions on LLM predictions.
In the dataset of \cite{Yasseri2022}, there are also 8 cases whose $H_1$ hints do not directly relevant to the answers of the corresponding questions. 
For example, in the third row of Figure~\ref{fig:table1}, the question is asking the number of upvotes to a post, but it’s $H_1$ hint of “Number of subscribers to the subreddit: 25.200” is not directly relevant to the question. 
We try to use these cases to test the robustness of bias effect (refer to Figure~\ref{fig:table4}).

\subsection{Experiment Design}
To evaluate the anchoring effects of LLMs, we also need to prompt a LLM $M$ to answer question $Q$ under different anchoring hints.
For each question $Q$, we design three experiments: control $C$, treatment $A$, and treatment $B$. 
In a control experiment, $Q$’s first hint $H_1$ is shown to LLM $M$ along with the question $Q$ as a part of the prompt string.
In treatment experiment $A$, besides the question $Q$, the anchoring hints $H_1$ and $H_2$ are both shown to LLM $M$.
Similarly, hints $H_1$ and $H_3$ are shown to $M$ in treatment experiment $B$.
Through systematically manipulating the visibility of the anchoring hints $H_2$ and $H_3$ to LLM $M$, we can examine the influence of anchoring hints on the answers from LLM $M$. 

The results generated by LLM are often sensitive to answer our questions.
In this study, we follow the CO-STAR format \cite{Teo2023} to compose our prompt and guide the LLM towards the desired outcome, which often consists of Context, Objective (i.e., the description of a task), Style \& Tone, Audience, and Response (i.e., output format).
Because our task is relatively simple, it only needs LLMs to output a numerical value rather than a long passage. 
So, we ignore the Style \& Tone and Audience parts because these parts are mainly used to specify the desired writing style and emotional tone for your LLM’s response.
Our prompt mainly consists of a context for background details including role playing, an objective of task description, and a specification of output format.
Our basic prompt template is presented in Appendix A.
As the provided hints often do not give enough information for an accurate answer to the question, we found that, if we simply ask a strong LLM (i.e., GPT4) to predict a value to answer our questions, the model sometimes directly refuses to give any answers.
Instead, it replies to us with “no enough information for prediction”. 
In the prompt template, we ask the LLM to generate an educated guess to answer our question rather than a prediction, which can mitigate the refuse-answering issue.
We ask the LLM to generate an answer with a numerical value and its unit using a JSON like format. 
In our experiments, all Control and Treatment $B$ \& $C$ share the same prompt template through replacing the placeholders of “user\_question”, “hint\_1”, and “hint\_2” at runtime. 
For each question $Q$, we replace “user\_question”, “hint\_1” by $Q$ and $H_1$. 
In a Control experiment, “hint\_2” is empty, and it is replaced by $H_2$ and $H_3$ respectively for different Treatment experiments.

In our study, we evaluate anchoring bias on three LLMs: GPT4o, GPT4, and GPT 3.5 Turbo with different settings. 
For each question, we ask every LLM model using a prompt composed based on the prompt template, collect and parse the answer from the LLM model to extract a numerical result. 
At the answer parsing step, we convert datetime values in the format of "HH:MM” or "MM:SS” to a decimal value in the corresponding minimal unit, because decimal values are easy to be used in statistical analysis.
For example, “2:00 with format of HH:MM” will be converted to a decimal value of "120" with a unit of "minute". 
Similarly, "2:00 of MM:SS” will be converted to "120" with a unit of "second".
For each question, we run 30 times on every language model to obtain enough data for statistical analysis. 
In the experiments, we set the temperature value as 0.8. 
\section{Result Analysis}
\label{sec:result_analysis}
\subsection{The Effect of Hint 1}
\label{subsec:effect}
\begin{figure*}
    \centering
    \includegraphics[width=\linewidth]{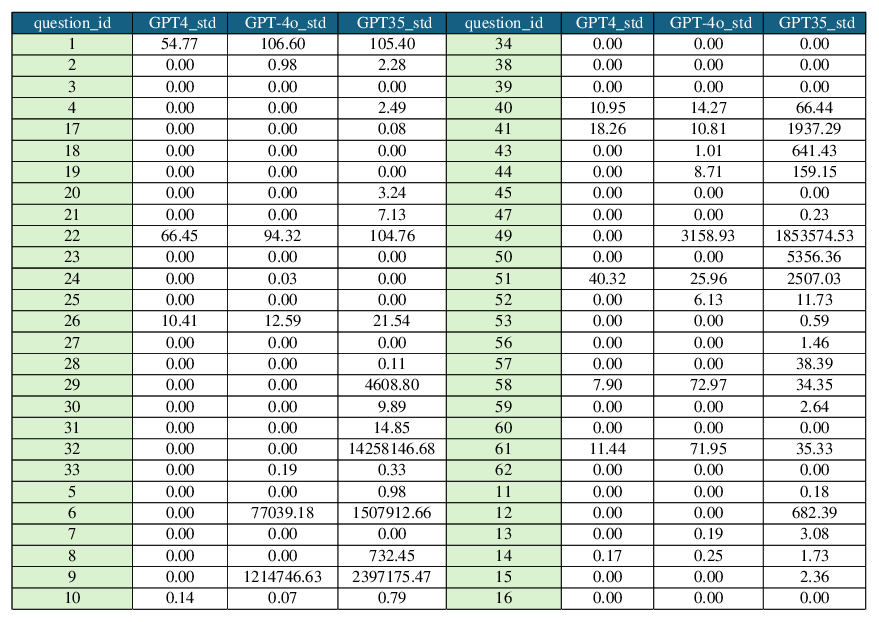}
    \caption{Standard deviations of answers from different models given fact hint $H_1$}
    \label{fig:table2}
\end{figure*}

Figure \ref{fig:table2} shows the standard variation values of responses from different LLM models (i.e., GPT4, GPT4o, and GPT3.5) on Control experiments.
From the figure, we found that many LLM responses of strong models (GPT4 and GPT4o) only have very small variations although a relatively large temperature (i.e. 0.8) is used in the experiments. 
For example, in 78 experiments (out of 108 total experiments) from GPT4 and GPT4o, the standard variation values are almost zero, which means the LLMs precisely repeat the same numerical answer 30 times.
At the same time, for a weaker model GPT3.5, only one fourth of experiments have zero variation values. 
A reasonable explanation for this is that a weak model (i.e., GPT3.5) is often not confident about its output.

\begin{figure*}
    \centering
    \includegraphics[width=\linewidth]{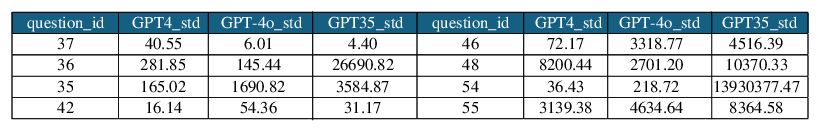}
    \caption{Standard deviations of answers from different models given irrelevant hints}
    \label{fig:table3}
\end{figure*}

As mentioned in section~\ref{sec:our_approach}, there are eight samples in the dataset, where each sample’s $H_1$ is irrelevant to its question. 
We can use these samples to check whether the models only blindly copy the number from hint $H_1$ as their outputs without understanding the user question and the content of hint $H_1$,
Figure \ref{fig:table3} shows the standard variation values of the responses from all LLMs. 
We can see that all these 8 cases have large variation values on both GPT4 and GPT4o, which are quite different from other samples in Figure~\ref{fig:table2}.
It means that LLMs do answer the questions based on the content understanding of hint $H_1$, and they cannot give confident responses when they are given an irrelevant hint $H_1$.

\subsection{Anchoring Effect Evaluation}
To evaluate the anchoring bias effect, we need to check whether the answers of LLMs are significantly influenced by hint $H_2$ or $H_3$ in the experiments. 
If anchoring effect exists, we are able to expect those answers with a high hint figure to obtain higher answer values that those who were provided with low hint figures. 
In other words, if the answers of a LLM under $H_2$ and $H_3$ are significantly different and the sign of the difference between them is the same as that of the difference between the original $H_2$ and $H_3$ values, we can claim that the answers of the LLM are biased by the anchors $H_2$ and $H_3$. 

\begin{figure*}
    \centering
    \includegraphics[width=\linewidth]{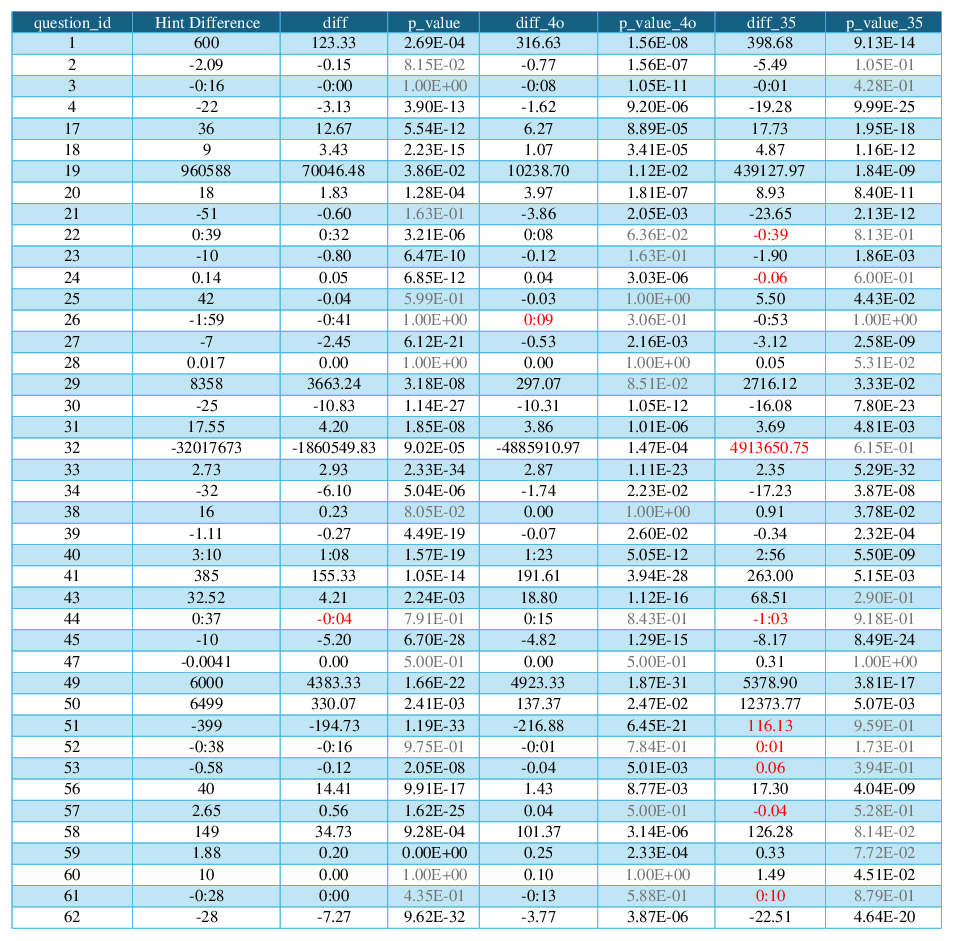}
    \caption{T-test results of biased answers on fact hints}
    \label{fig:table4}
\end{figure*}

\begin{figure*}
    \centering
    \includegraphics[width=\linewidth]{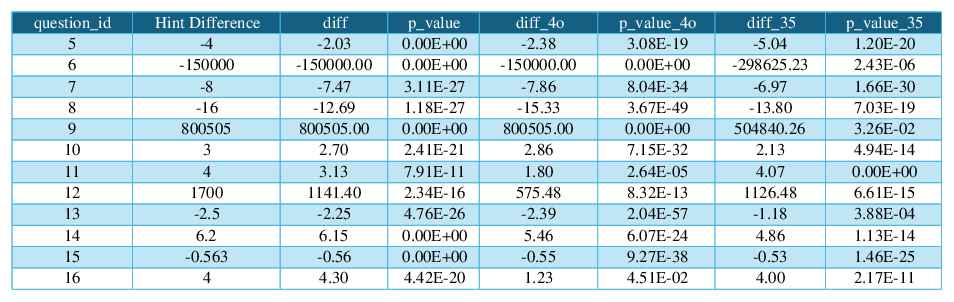}
    \caption{T-test results of biased answers on expert hints}
    \label{fig:table5}
\end{figure*}

In our experiments, for each question, we obtain two value sets from every language model: one set contains 30 answers for Treatment $A$ (i.e. answer set $A$), and the other contains 30 for Treatment $B$ (i.e. answer set $B$). 
We evaluate the bias effect by estimating the average difference between two underlined distributions of answer in set $A$ and set $B$ and measure the confidence of this average difference with $p$-value using t-test analysis. 
Figure~\ref{fig:table4} and \ref{fig:table5} show the $t$-$test$ results on different LLMs’ responses to questions with different hint types of fact anchoring and expert opinion anchoring, respectively.
Here, the column “Hint Difference” shows the difference between the original values in two anchoring hints $H_2$ and $H_3$ for each question (e.g., $H_3 - H_2$). 
The columns “diff”, “diff\_4o”, and “diff\_35” are the average difference values between the two distributions from treatment experiment $A$ and $B$ (e.g., $(B-A)$) on the GPT-4 model, GPT-4o, and GPT3.5 respectively. And, “$p$\_value”, “$p$\_value\_4o”,  and “$p$\_value\_35” are their corresponding t-test confidence values.
A smaller p-value (i.e., less than 0.05) often means a higher confidence on the hypothesis that two distributions are different.

\subsubsection{Observation 1. Stronger models are consistently biased by anchor hints.}
From Figure~\ref{fig:table4} and \ref{fig:table5}, we can see that the answers to many questions are biased by the anchoring hints. 
Specifically, most distribution differences (i.e., “diff”, “diff\_4o”, and “diff\_35” values in the figure) share the same signs with their corresponding “Hint Difference” values.
For example, there are only 11 exception cases (highlighted with red color in the tables) among the total 162 distribution difference values, whose signs are not consistent with the signs of their corresponding "Hint Difference" values. 
It provides a strong evidence to the anchoring effect of LLMs, although a few $p$-values show that the corresponding differences are not statistically significant enough. 
It also shows that the answers from stronger models (i.e., GPT-4 and GPT-4o) are more consistently influenced by anchoring bias. 
On the contrary, the anchoring effects are weaker on GPT3.5.
Among the 11 exceptions, 9 of them come from the answers of GPT3.5. 
We think it is reasonable because a weaker model may introduce a lot of randomness in answer generation that leads to a high fluctuation of the results, while stronger models have more stable answer values. 

\subsubsection{Observation 2. Anchoring effects on time-related questions not significant.}
By looking into cases in Figure~\ref{fig:table4}, we have an interesting observation that the questions with time-period answers are not easy to be biased. 
For example, in Figure~\ref{fig:table2}, we have 7 questions ($no.$3, $no.$22, $no.$40, $no.$26, $no.$44,  $no.$52, $no.$61) expecting time-period answers, which are about 17\% (7 of 42) of the questions, but they bring about 54\% (6 of 11) out of total exceptions. 
Our hypothesis is that the capability of LLM’s time-period prediction is slightly worse than its capability on normal numbers.

\subsubsection{Observation 3. LLMs are much easier to be biased by expert anchors.}
Figure~\ref{fig:table5} shows that the answers from GPT models are biased by expert anchors with high confidence. 
For example, there is no exception in the answers of 12 questions whose anchor hints come from a PTF (Play The Future) expert. 
Furthermore, their corresponding $p$-values are also very small, which means that the system has a very high level of confidence. 
In other words, the results show that all models including strong and weak models tend to strongly adherent to the expert hints from PTF. 

\subsubsection{Observation 4. Anchoring effects are not significant when given irrelevant hints.}
\begin{figure*}
    \centering
    \includegraphics[width=\linewidth]{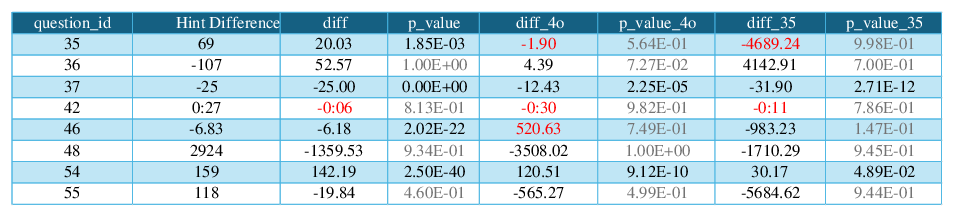}
    \caption{T-test results of biased answers when $H_1$ is irrelevant to the question}
    \label{fig:table6}
\end{figure*}

Figure~\ref{fig:table6} shows that all GPT models do not give consistent behaviors for questions whose $H_1$ hint does not provide relevant information to the question. 
It seems that the randomness of answer values is overwhelming the potential anchoring effect when no enough information is provided in hint $H_1$ to LLMs for question answering.

\subsection{Discussion}
It is interesting to know whether LLMs perform similarly to human users. We compared our results from LLMs to the results from the human user study in \cite{Yasseri2022}.

\textit{First}, in the human user study on the same dataset from \cite{Yasseri2022}, the human responses often have non-zero variations and the authors noticed loyal users in the control group make predictions that resemble the median predictions more closely than casual users.
In our experiments, we have a similar observation (refer to \ref{subsec:effect}) that stronger models (i.e., GPT4, GPT4o) have smaller variations than the weaker model (i.e., GPT3.5).
In other words, whether it's a human or a model, there is a common rule: as the ability improves, the variance of prediction becomes smaller and the accuracy becomes higher.

\textit{Second}, the average anchoring index (AI \cite{Jacwitz1995}) of GPT4 is about 0.45 in our experiment, it is different from 0.61 in the human study \cite{Yasseri2022}.
In other words, the level of influence created by anchoring hints on LLMs is smaller than that of humans. 
It seems that human is easier to be influenced by anchoring hints.
\[
AI = \frac{\text{median}_{\text{high anchor}} - \text{median}_{\text{low anchor}}}{\text{high anchor} - \text{low anchor}}
\]

\textit{Third}, although the analysis method is different from that in the human user study \cite{Yasseri2022}, from Figure~\ref{fig:table4}, we also can have the same conclusion that a larger stimulus value (refer to the Hint difference column that is the value difference between hints $H_2$ and $H_3$) often leads to a larger anchoring response.

\textit{Fourth}, \cite{Yasseri2022} reports that human users’ responses to the questions containing PTF-prediction values are consistently larger than that to the standard questions.  
In our experiment, the results also show that the responses of LLMs are strongly adherent to the anchored hints of PTF-prediction.
It seems that both human users and LLMs are easy to influenced by expert-opinions.

\section{Mitigation Strategies}
\label{sec:mitigate_strategy}
As cognitive bias may largely influence the correctness of decision making, it is important to know whether we can mitigate the anchoring bias of LLMs during inference. 
In this section, we list several potential strategies and test whether they can mitigate anchoring bias effectively.

Here is a list of strategies we try to study in this paper:
\begin{itemize}
    \item \textbf{Chain-of-Thought (CoT)}: it is well known that chain-of-thought can significantly improve reasoning accuracies of LLM models~\cite{Wei2022}. 
    It is reasonable to ask whether CoT can mitigate anchoring bias of LLMs.
    \item \textbf{Thoughts of Principles}: some previous studies~\cite{Zheng2024} have shown if we ask LLM to generate principles before reasoning, the overall reasoning accuracy can be boosted. 
    In this section, we test whether the thoughts-of-principle strategy (denoted as PoT in our experiments) can reduce the influence of anchoring bias.
    \item \textbf{Ignore Anchor Hint}: as we know, the bias is caused by anchor hints. 
    In this strategy, similar to \cite{Jeremy2024}, we try to mitigate the anchoring effect by explicitly asking LLMs to ignore the anchor hints. 
    \item \textbf{Reflection}: reflection is another widely used prompting strategy that can boost reasoning accuracy of LLMs. 
    It lets a LLM mimic human’s reflective thinking to improve reasoning accuracy through self-correcting its reasoning steps iteratively. 
    In this section, we also check whether we can use reflection prompting to mitigate anchoring bias.
    \item \textbf{Both-Anchor}: In human decision-making practices, a practical strategy to mitigate the impact of anchoring bias is to gather information from as many angles as possible to serve decision-making.
    This prevents our cognition from being anchored to individual pieces of information, thereby allowing for a more comprehensive judgment \cite{Krockow2019}.
    To simulate multi-angle information, in this study, we attempted to include both $H_2$ and $H_3$ in the prompts to LLM to simulate a situation of more comprehensive prompt information.
\end{itemize}

In our experiments, we modify our prompt template to integrate these mitigation strategies. 
For example, prompt $b$ in Appendix is our prompt template for the “ignore” strategy, where we ask the LLM to ignore the anchor hint by adding an instruction “The hint part contains an answer from a PTF expert and please **ignore** it when you are answering the question” into the prompt.
 Then, we test these prompts using GPT-4 and GPT-4o on our “expert” anchoring questions. 
We use the “expert” anchoring questions because the anchoring effects are very significant for these questions.
\cite{Simmons2010}.
\begin{figure*}
    \centering
    \includegraphics[width=\linewidth]{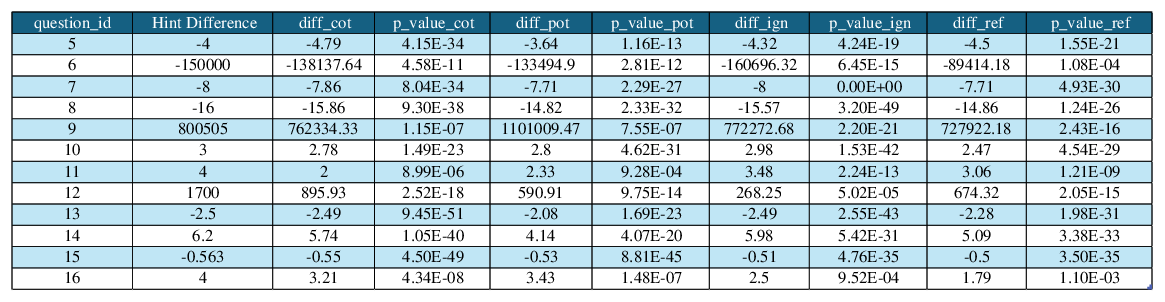}
    \caption{T-test results of biased answers under different mitigation strategies with GPT-4o. }
    \label{fig:table7}
\end{figure*}

\begin{figure*}
    \centering
    \includegraphics[width=\linewidth]{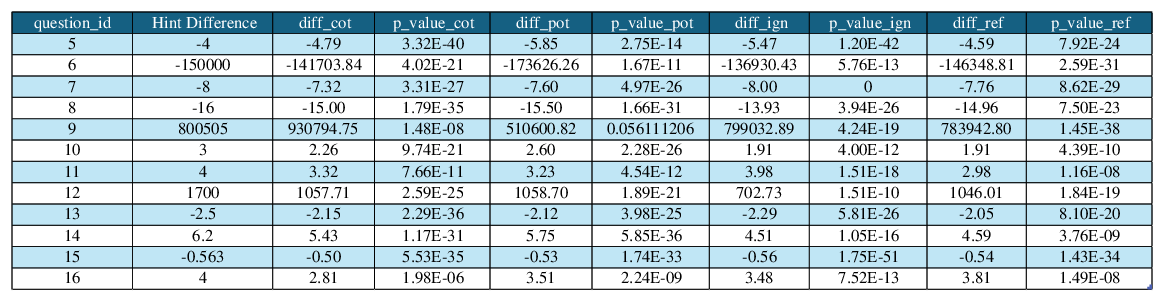}
    \caption{T-test results of biased answers under different mitigation strategies with GPT-4. }
    \label{fig:table8}
\end{figure*}

Figure \ref{fig:table7} and \ref{fig:table8} are our results obtained from GPT-4o and GPT-4 respectively.
Here, diff\_cot, diff\_pot, diff\_ign, diff\_ref, and p\_value\_cot, p\_value\_pot, p\_value\_ign, p\_value \_ref are the differences and p-values under different strategies including Chain of Thought (CoT), Principle of Thought (PoT), Ignore Strategy (Ign), and Refection Strategy (Ref) respectively.
From the figures, we can see that the answers from models with hints of $H_2$ and $H_3$ are still consistently influenced by the anchoring hints, and their corresponding $p$-values are very small.
By comparing with Figure~\ref{fig:table5}, we do not find any significant improvements. 
It seems that all these strategies are not effective enough to mitigate the anchoring bias for these questions. 
Similarly, Simmons et.al. also observed that it is difficult to completely eliminate anchoring bias from their human experiments~\cite{Simmons2010}.

\begin{figure}
    \centering
    \includegraphics[width=\linewidth]{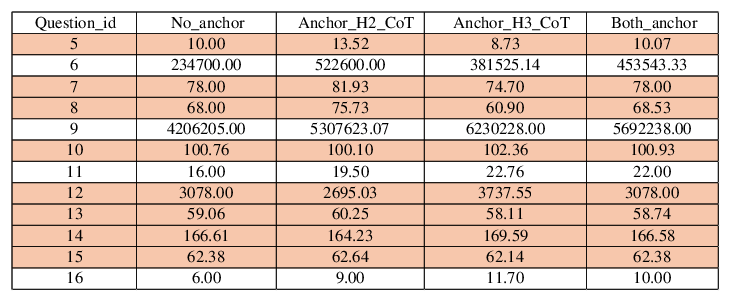}
    \caption{The average answer values obtained from GPT-4 from the CoT strategy, the Both-Anchor strategy, and without anchoring hints ($No\_anchor$ column). }
    \label{fig:table9}
\end{figure}

Different from other strategies, our Both-Anchor strategy does not try to directly eliminate anchoring effect. 
Instead, it includes both $H_2$ and $H_3$ in the prompt to prevent the LLMs from being anchored to an individual hint.
We conducted experiments on questions with expert hints using GPT-4, and obtained the results shown in Figure~\ref{fig:table9} (noted as "Both\_anchor" column). 
In the figure, we observed that when the values of the $H_2$ and $H_3$ prompts are on either side of the bias-free benchmark value (e.g., refer to the questions no.5, no.7, no.8, no.10, no.12, no.13, no.14, and no.15), the answers of GPT-4 under the Both-Anchor strategy are very close to the bias-free benchmark value.
On the contrary, the answers of GPT-4 to the remaining questions still deviate significantly from the benchmark value. 
This phenomenon suggests that the advice from human psychology \cite{Krockow2019} can be applicable to mitigate the impact of anchoring bias in LLMs in practical applications to certain extent, although it does not try to eliminate anchoring bias from a single anchoring hint. 
\section{Conclusion}
\label{sec:conclusion}
In conclusion, our study shows that anchoring bias is widely prevalent in large language models (LLMs) and poses a significant challenge to their reliability and fairness. 
Especially, strong models consistently show their vulnerability to the bias of anchoring effect. 
They are very easy to be influenced by “expert” opinions presented in the prompt, and we cannot correct their behavior even when we explicitly ask them not to follow the expert opinions.  
At the same time, we tried various strategies, including Chain-of-Thought, Thoughts of Principles, Ignoring Anchor Hints, and Reflection, to mitigate anchoring bias in LLMs. 
Despite the promise application of these strategies in many reasoning tasks, our empirical results on GPT-4 and GPT-4o indicate that none of these simple mitigating strategies can effectively reduce the anchoring bias in responses to "expert" anchoring questions. 
This finding underscores the complexity of cognitive biases in AI and highlights the need for further research to develop more robust techniques for bias mitigation.
All the Prompt templates and LLMs answers can be found from the URL: https://github.com/JiaxuLou/LLM\_Bias.
Furthermore, our study also show that,  although it is impossible to completely eliminate anchoring effects, in practice, one can collect hints from comprehensive angles to mitigate the impact of anchoring bias by preventing the LLMs from being anchored to individual pieces of information.

\section*{Acknowledgment}
I thank Prof. Taha Yasseri for his kind help of sharing me the questions, program, and human answers from their experiments \cite{Yasseri2022}. Our experiments are designed based on these questions.

\section{Appendix}
\subsection{Prompt template with anchoring hints}
\hspace{1pt}

<|im\_start|>System

You are a Copilot, a thoughtful and intelligent assistant to help answer a user question.
Beside the question description, there are at least one hint that may or may not be relevant to the user question. You must give an answer to the question using an educated guess based on the provided information.
Each answer is a number following a proper unit mentioned in the user question. 

Your answer **must** uses the follow format. 

\#\#\# Output format: 

\{

"number": ...,

"unit": ...

\}

<|im\_end|>

<|im\_start|>assistant

OK I'm an assistant and I'll be helping users to give an educated guess for each user question. 

<|im\_end|>

<|im\_start|>User

Good! Here is the question and hints.

\#\#\# Question

{{user\_question}}

\#\#\# Here are some hints

{{hint\_1}}

{{hint\_2}}

Please give your answer!

<|im\_end|>

<|im\_start|>assistant

OK, based on the provided information, I give the following answer.
The answer to the Question is:

\subsection{An Example of Prompt and Answer from GPT4}
\subsubsection{The Question}
\hspace{1pt}

<|im\_start|>System

You are a Copilot, a thoughtful and intelligent assistant to help answer a user question.
Beside the question description, there are at least one hint that may or may not be relevant to the user question. You must give an answer to the question using an educated guess based on the provided information.
Each answer is a number following a proper unit mentioned in the user question. 

Your answer **must** uses the follow format. 

\#\#\# Output format: 

\{

"number": ...,

"unit": ...

\}

<|im\_end|>

<|im\_start|>assistant

OK I'm an assistant and I'll be helping users to give an educated guess for each user question. 

<|im\_end|>

<|im\_start|>User

Good! Here is the question and hints.

\#\#\# Question

Apple's Worldwide Developers Conference will take place next week in San Jose where exciting new software and devices may be presented! What will the starting price be in US \$ of the device that is announced last during Apple’s WWDC 2018?

\#\#\# Here are some hints

Current starting price of Apple’s iPhone 8: 699 USD.

Current starting price of Apple’s Watch Series 3 (Cellular): 399 USD

Please give your answer!

<|im\_end|>

<|im\_start|>assistant

OK, based on the provided information, I give the following answer.
The answer to the Question is:

\subsubsection{Answer from GPT4}
\hspace{1pt}

\{

"number": 499,

"unit": "USD"

\}

\section{Prompt template for Mitigation strategy of "Ignore Anchor Hint" }
\hspace{1pt}

<|im\_start|>System

You are a Copilot, a thoughtful and intelligent assistant to help answer a user question.  
Beside the question description, there are at least one hints that may or may not be relevant to the user question. 
\textbf{The hint part contains an answer from a PTF expert and please **ignore** it when you are answering the question.}
You must give an answer to the question using an educated guess based on the provided information. 
Please first illustrate your principles how you obtain the answer and then give your answer.
Each answer is a number following a proper unit mentioned in the user question. 

Your answer [must] uses the follow format. 

\#\#\# Output format:

\{

   "principles": ...,
   
   "number": ...,
   
   "unit": ...
   
\}

<|im\_end|>

<|im\_start|>assistant

OK I'm an assistant and I'll be helping users to give an educated guess for each user question. 

<|im\_end|>

<|im\_start|>User

Good! Here is the question and hints.

\#\#\# Question

{{user\_question}}

\#\#\# Hints

{{hint\_1}}

{{hint\_2}}

<|im\_end|>

<|im\_start|>assistant

\textbf{OK, based on the provided information, I give the answer step by step.}

\end{document}